%% file: root.tex
\documentclass[conference]{IEEEtran}
\IEEEoverridecommandlockouts
\usepackage{cite}
\usepackage{amsmath,amssymb,amsfonts}
\usepackage{algorithmic}
\usepackage{graphicx}
\usepackage{textcomp}
\usepackage{xcolor}
\usepackage{xspace}
\usepackage{subcaption}
\def\BibTeX{{\rm B\kern-.05em{\sc i\kern-.025em b}\kern-.08em
    T\kern-.1667em\lower.7ex\hbox{E}\kern-.125emX}}
    
\usepackage{multirow}
\usepackage{hyperref}
    
\newcommand*{\aruco}{ArUco\@\xspace}

\newcommand{\vspacebeforesubcaption}[0]{\vspace{-5mm}}
\newcommand{\vspacebetweensubfigures}[0]{\vspace{2mm}}
\newcommand{\vspaceafterfigure}[0]{\vspace{-4mm}}  
\begin{document}


\title{BURG-Toolkit: Robot Grasping Experiments in Simulation and the Real World
    \thanks{
    The work is funded by EPSRC (grant no. EP/S032487/1) and FWF (grant no. I3967-N30) through CHIST-ERA project BURG, and by EC grant agreement no.~101017089, project TraceBot.
    }
}

\author{Martin Rudorfer$^{1}$, Markus Suchi$^{2}$, Mohan Sridharan$^{1}$, Markus Vincze$^{2}$, Ale\v{s} Leonardis$^{1}$%
\thanks{$^{1}$M.~Rudorfer, M.~Sridharan and A.~Leonardis are with the School of Computer Science at the University of Birmingham, UK.
        {\tt\small \{m.rudorfer, m.sridharan, a.leonardis\}@bham.ac.uk}}%
\thanks{$^{2}$M.~Suchi and M.~Vincze are with the Automation and Control Institute, TU Wien, Austria.
        {\tt\small \{suchi, vincze\}@acin.tuwien.ac.at}}%
}

\maketitle

\begin{abstract}
This paper presents BURG-Toolkit, a set of open-source tools for Benchmarking and Understanding Robotic Grasping. Our tools allow researchers to: (1)~create virtual scenes for generating training data and performing grasping in simulation; (2)~recreate the scene by arranging the corresponding objects accurately in the physical world for real robot experiments, supporting an analysis of the sim-to-real gap; and (3)~share the scenes with other researchers to foster comparability and reproducibility of experimental results. We explain how to use our tools by describing some potential use cases. We further provide proof-of-concept experimental results quantifying the sim-to-real gap for robot grasping in some example scenes. The tools are available at: \url{https://mrudorfer.github.io/burg-toolkit/}
\end{abstract}


\section{Introduction}
\input{01-introduction}


\section{BURG-Toolkit}
\input{02-burg-tools}

\section{Experiments}
\input{03-experiments}

\section{Outlook}
\input{04-conclusion}

\bibliographystyle{IEEEtran}
\bibliography{root}

\end{document}

%% file: 01-introduction.tex
Grasping and manipulation of novel objects in complex, cluttered environments is an open problem. A common component of many approaches to this problem is to estimate and rank potential 6-DOF grasp poses based on an input RGBD image and to execute the highest-ranked grasp candidate using traditional motion planning. In recent years, many data-driven methods have been proposed for predicting grasps for isolated objects~\cite{GPD_2017,PointNetGPD_2019,6DOF_GraspNet_2019,GPNet_2020}
and for objects in cluttered scenes~\cite{VolumetricGraspingNetwork_2020,S4G_2020,GraspNet1Billion_2020,ContactGraspNet_2021,SemanticAndCollisionLearning4Grasping_2021}.
The vast majority of these methods are trained using purely synthetic data created by rendering images of 3D object models. The required grasp annotations are gathered by performing antipodal or approach-based grasp sampling~\cite{BillionWaysToGrasp_2019} on 3D object models and evaluating these grasps either with analytic metrics such as force closure~\cite{ForceClosure_1988} and grasp wrench space~\cite{FerrariCanny_1992} or by executing them in a physics simulation such as PyBullet~\cite{PyBullet} or MuJoCo~\cite{MuJoCo}. To demonstrate performance in the real world, all of the mentioned works also present real robot grasping experiments.

\begin{figure}[tb]
	\centering
    \begin{subfigure}[t]{0.48\columnwidth}
       \centering
       \includegraphics[width=\textwidth]{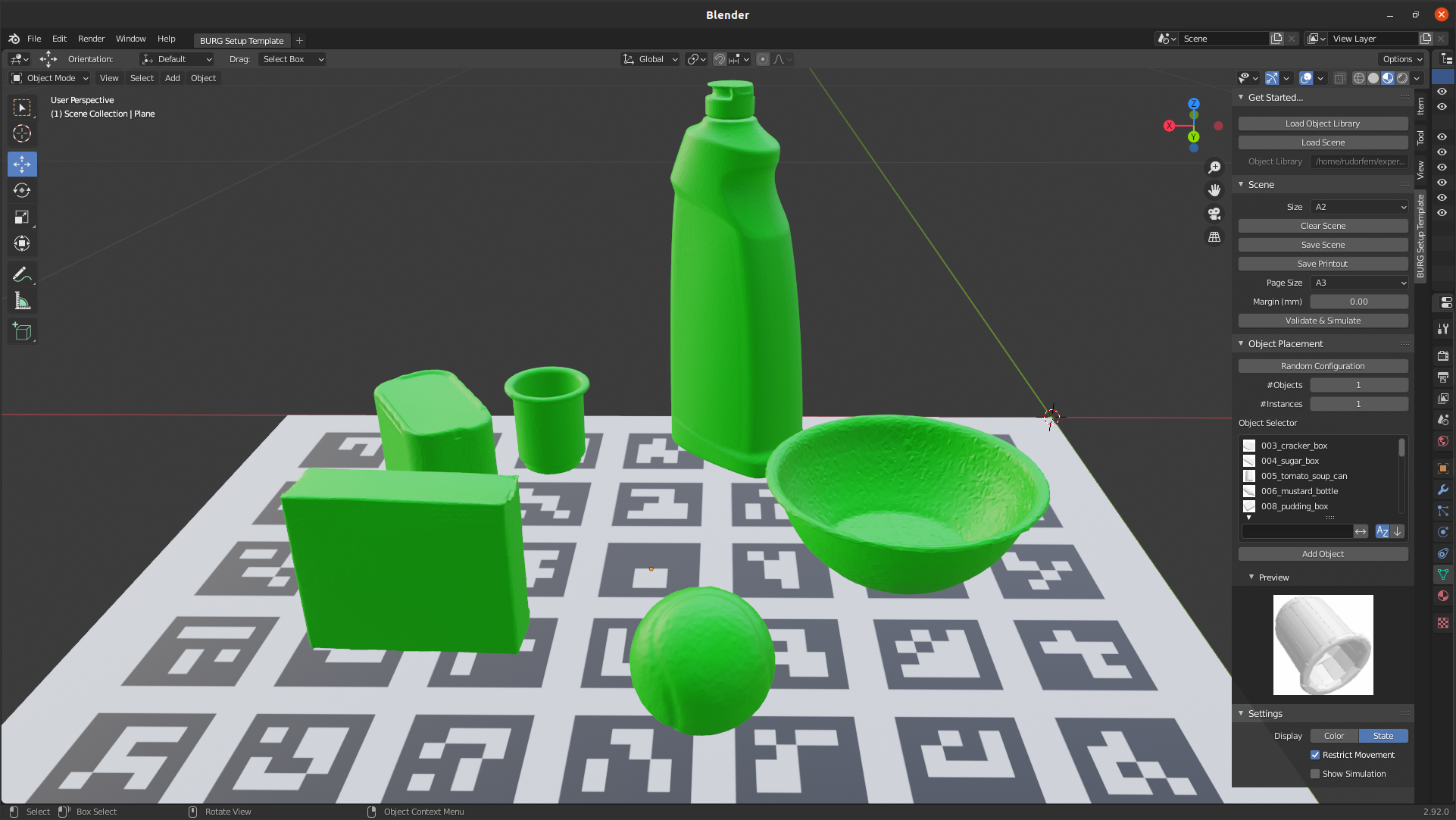}
       \vspacebeforesubcaption
       \caption{Scene creation in SetupTool}
  	\end{subfigure}
  	\begin{subfigure}[t]{0.48\columnwidth}
       \centering
       \includegraphics[width=\textwidth]{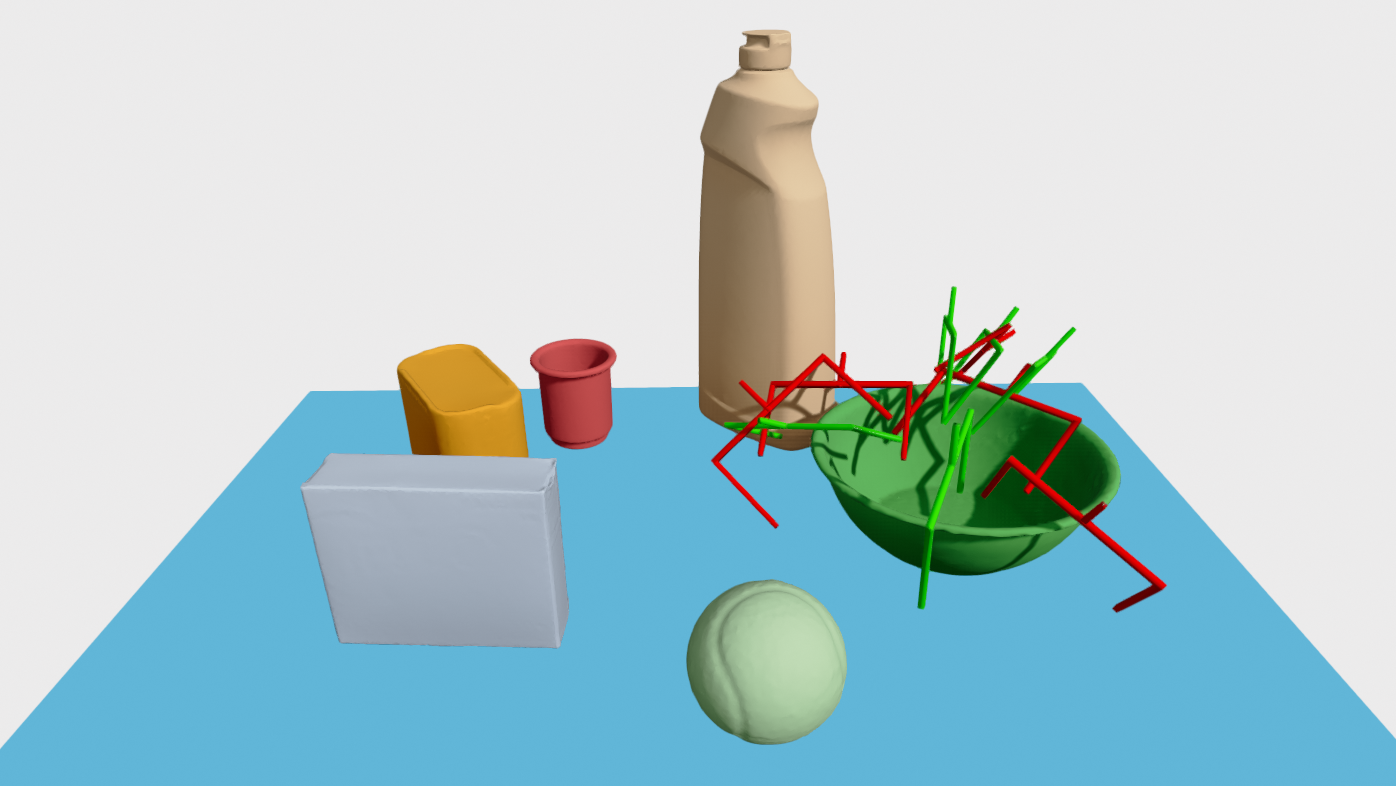}
       \vspacebeforesubcaption
       \caption{Sampling grasps for the bowl}
  	\end{subfigure}
  	
  	\vspacebetweensubfigures
  	\begin{subfigure}[t]{0.48\columnwidth}
       \centering
       \includegraphics[width=\textwidth]{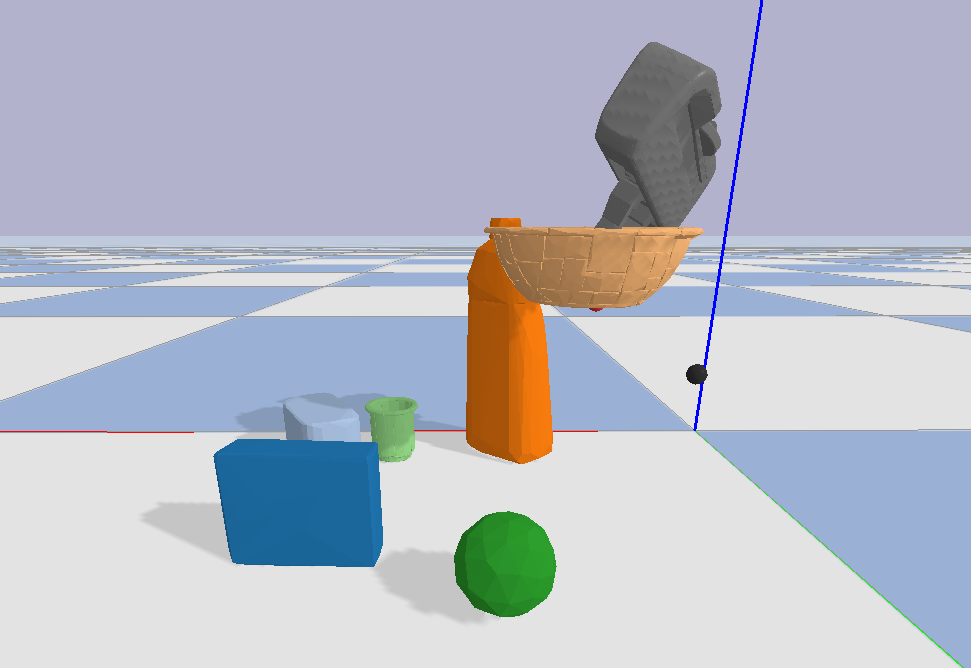}
       \vspacebeforesubcaption
       \caption{Simulation-based grasp execution}
  	\end{subfigure}
  	\begin{subfigure}[t]{0.48\columnwidth}
       \centering
       \includegraphics[width=\textwidth]{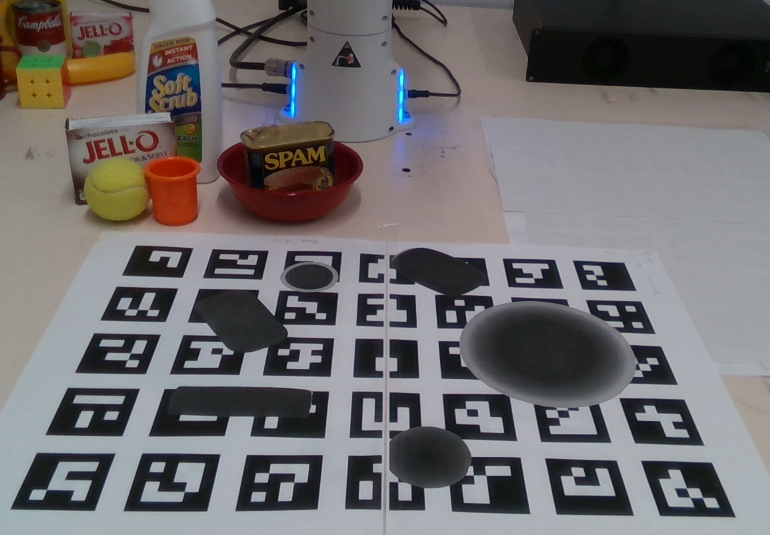}
       \vspacebeforesubcaption
       \caption{Printout with placement indications}
  	\end{subfigure}
  	
  	\vspacebetweensubfigures
  	\begin{subfigure}[t]{0.48\columnwidth}
       \centering
       \includegraphics[width=\textwidth]{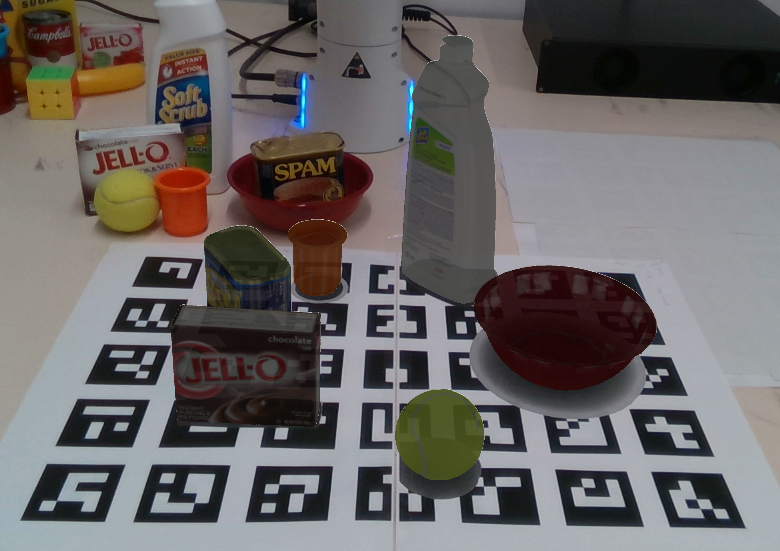}
       \vspacebeforesubcaption
       \caption{Printout with AR projections}
  	\end{subfigure}
  	\begin{subfigure}[t]{0.48\columnwidth}
       \centering
       \includegraphics[width=\textwidth]{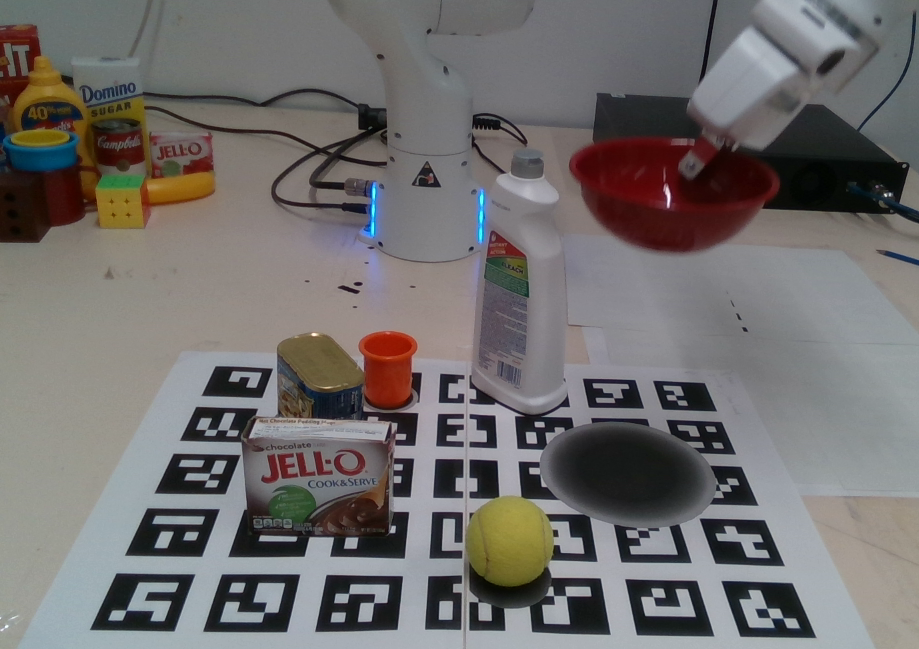}
       \vspacebeforesubcaption
       \caption{Real-world grasp execution}
  	\end{subfigure}
  	\caption{
  	    Exemplary workflow: (a-c) After creating the virtual scene using our SetupTool, we can sample grasps and execute them in a physics-based simulation; (d) From the SetupTool we can export a printout with an \aruco marker board and the object height maps indicating the placements; (e) To further facilitate the placement, our SceneVisualiser projects the objects into the camera image for the user to make sure the real objects align visually; and (f) Finally, the scene is set up for real robot experiments.
  	}
  	\vspaceafterfigure
  	\label{fig:workflow_example}
\end{figure}

The results from synthetic evaluation and real-world experiments are unfortunately not comparable across the many existing methods. Researchers often tailor synthetic datasets to their specific needs and potentially generate auxiliary annotations. Recently, a large-scale grasp dataset has been proposed~\cite{ACRONYM_2020}, but it is designed and used more for training than evaluation and is yet to be taken up by the wider community. 
Real-world experiments are often conducted with different object sets by different people. Some researchers enhance reproducibility by using objects from the established YCB object set~\cite{YCB}, yet the specific scene compositions are often unknown or random, making it difficult to reproduce the experimental conditions in other labs. Improving the reproducibility of experiments will have a strong positive impact on research progress~\cite{ReproducibleRobotics_2017}.
GRASPA~\cite{GRASPA_2020} suggested the use of printable templates to determine scene composition and shared such templates for a small number of benchmark scenes. We build on this idea and go one step further by providing tools to create, share, and reproduce the exact setup of the objects for simulation-based and real robot evaluation.

The grasping methods trained in simulation are generally subject to two different domain gaps: the gap between rendered and real images, which is considered e.g. in~\cite{DomainAdaptationGrasping_2018,RetinaGAN_2021}, and the gap between simulated and real grasps.
The physics-based grasp simulations usually make simplifying assumptions, e.g., that objects have uniform density and identical friction properties~\cite{6DOF_GraspNet_2019,GPNet_2020} or that the effects of gravity can be disregarded~\cite{ACRONYM_2020}. While these choices are reasonable to keep computational demands tractable, the effects of these choices on real-world grasp trials are rarely investigated. We believe that systematically examining the differences between simulated and real-world grasp trials will provide valuable insights for improving the simulation environments and thereby reducing the sim-to-real gap. To this end, we seek to enable researchers to exactly reproduce their synthetic scenes for real robot experiments. This will help validate and improve the simulation environment and offer the opportunity to capture real-world images with grasp annotations.

This paper introduces BURG-Toolkit, a set of open-source tools for \textbf{B}enchmarking and \textbf{U}nderstanding \textbf{R}obotic \textbf{G}rasping. These tools enable researchers to setup scenes for simulation and the real world, and to share them with the community. We present use cases and describe results of a proof-of-concept experiment investigating the differences between simulation-based grasping and grasping with a real robot.


%% file: 02-burg-tools.tex
\begin{figure}[tb]
   \centering
   \includegraphics[width=0.9\columnwidth]{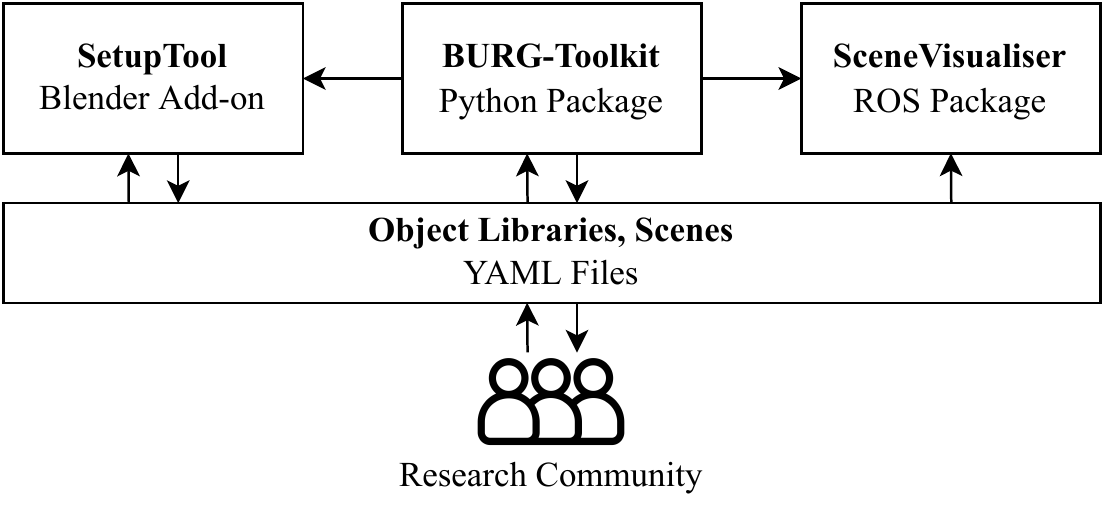}
   \vspace{-0.7em}
   \caption{Overview of the BURG-Toolkit, a Python package that provides functionality directly or as a backbone for the SetupTool and the SceneVisualiser.  Object libraries and scenes are stored in YAML files that can be shared with the research community.  }
   \label{fig:burg-tools-overview}
\end{figure}

The core of the \textit{BURG-Toolkit} is a Python package of the same name.
It contains data structures and methods implementing various capabilities, including a grasp simulator based on PyBullet~\cite{PyBullet}.
The Blender-based \textit{SetupTool} offers an intuitive user interface for the arrangement of 3D object models to form desired scenes.
We can use these scenes to perform simulation-based grasping trials.
In addition, we can generate a printout for each scene which indicates the object placements and simultaneously supports projecting the objects into the real scene using an Augmented Reality \textit{SceneVisualiser}; this is also provided as a package for ROS~\cite{ROS} so that it can be easily integrated with existing robot systems.
The printout and projection can be used to arrange the physical objects with high accuracy and perform real-world experiments that match the simulated grasping trials.
Fig.~\ref{fig:workflow_example} illustrates the workflow for using the BURG-Toolkit and Fig.~\ref{fig:burg-tools-overview} shows how our tools are organised. 
The SetupTool and the SceneVisualiser rely on the BURG-Toolkit as backend.
The generated scene data is stored in human-readable YAML files to allow easy editing and sharing with other researchers. We describe our tools further in the following subsections.

\subsection{Object Library and Scene Creation}
All target objects are organised in an object library. For each object, the user must provide an identifier, a mesh file, the mass, and optional properties like friction coefficient or scale factor. Based on this information, the BURG-Toolkit creates \emph{VHACD} and \emph{URDF} files for use in simulation and determines the stable poses of each object.
The stable poses are first computed based on the mesh and then validated in simulation.
All properties and file paths are stored in an \emph{object\_library.yaml} file which can be edited and shared with the referenced files.

\begin{figure}[tb]
    \centering
  	\begin{subfigure}[t]{0.48\columnwidth}
       \centering
       \includegraphics[width=\textwidth]{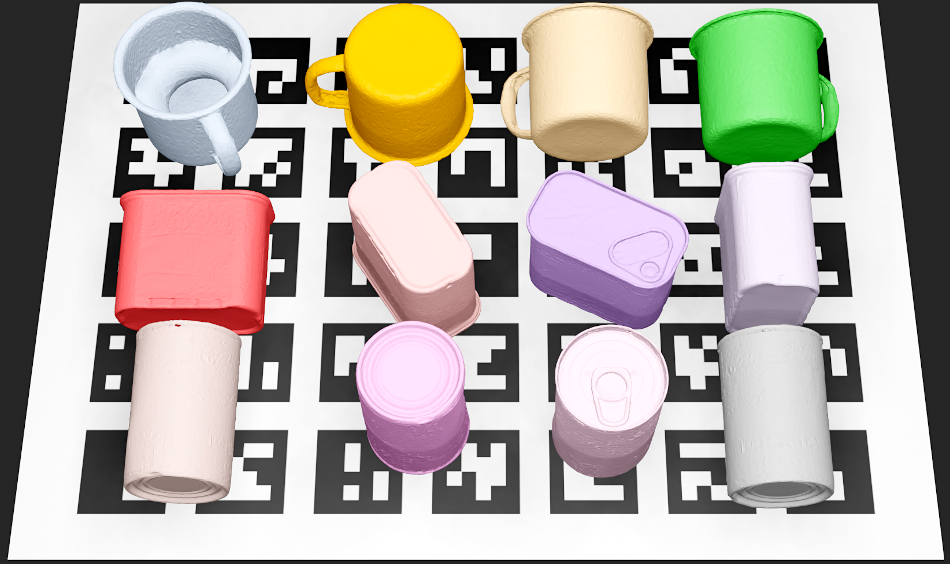}
       \vspacebeforesubcaption
       \caption{Stable poses of exemplary objects}
  	\end{subfigure}
  	\begin{subfigure}[t]{0.48\columnwidth}
       \centering
       \includegraphics[width=\textwidth]{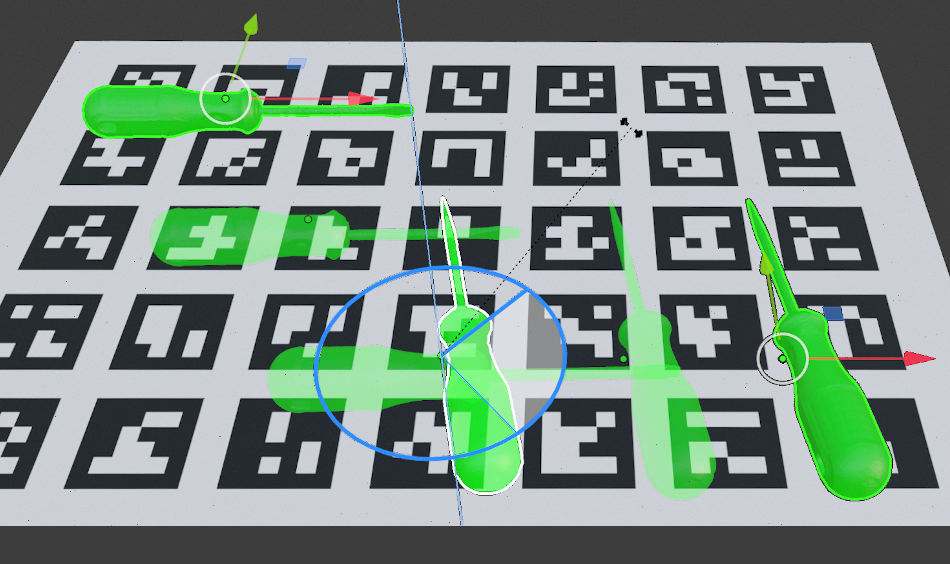}
       \vspacebeforesubcaption
       \caption{Manipulation in stable pose}
  	\end{subfigure}
  	
  	\vspacebetweensubfigures
    \begin{subfigure}[t]{0.48\columnwidth}
       \centering
       \includegraphics[width=\textwidth]{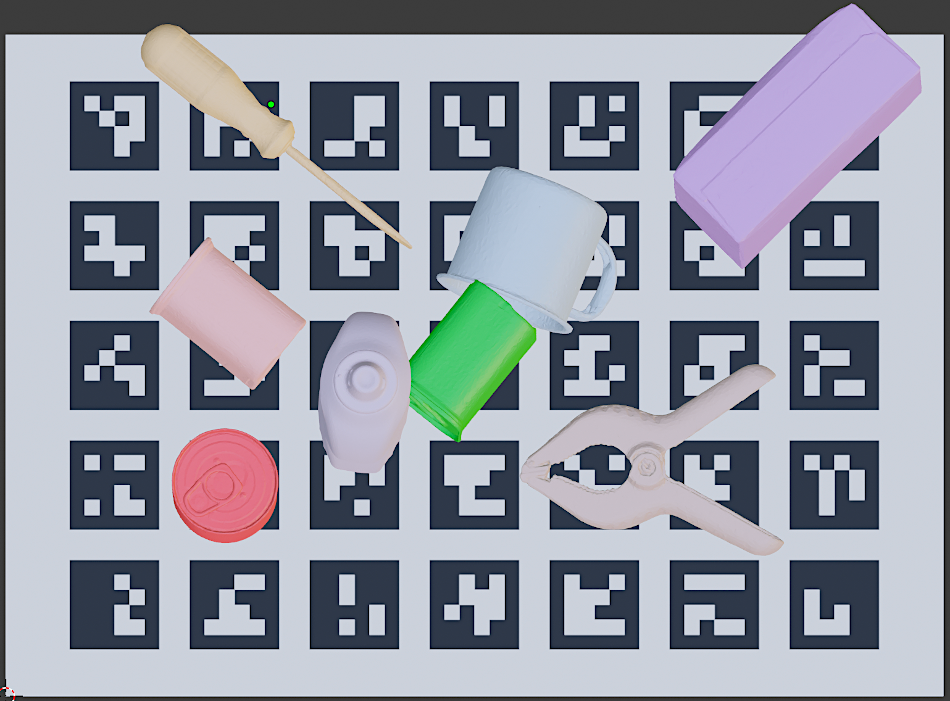}
       \vspacebeforesubcaption
       \caption{Scene composition}
  	\end{subfigure}
  	\begin{subfigure}[t]{0.48\columnwidth}
       \centering
       \includegraphics[width=\textwidth]{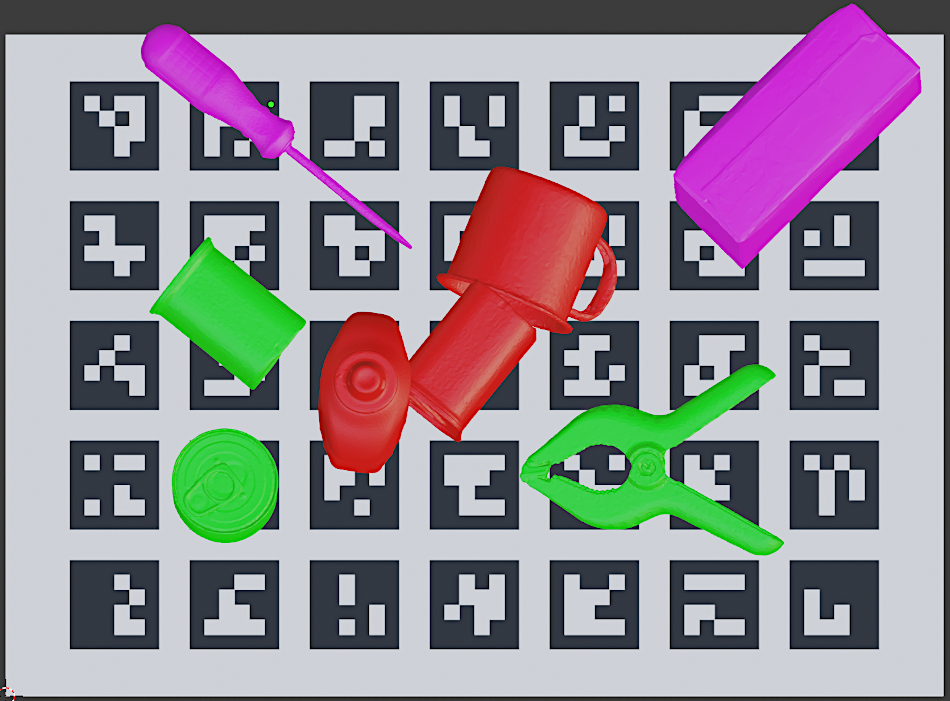}
       \vspacebeforesubcaption
       \caption{Status colouring after validation}
  	\end{subfigure}
  	\caption{Illustration of SetupTool's key features: (a) stable poses of three objects; (b) GUI options with movement restrictions to ensure stable pose during object manipulation; (c)~scene composition; and (d) post validation status is colour-coded as green: ok, red: collision, pink: out of bounds.}
    \vspaceafterfigure
  	\label{fig:toolkit_features}
\end{figure}

The objects in the library can be arranged in scenes with our SetupTool. This is realised as an add-on for the open-source 3D creation suite Blender~\cite{Blender} and benefits from Blender's well-established and customisable user interface for viewing and manipulating 3D scenes. An overview of important functions is given in Table~\ref{tab:setuptool-functions} and key features are highlighted in Fig.~\ref{fig:toolkit_features}. When adding objects to a scene, we can choose from the pre-computed stable poses and restrict movement to translation in~x/y and rotation around~z in order to keep the object stable. If more customisation is required, the user can move objects freely instead.
Scene validation enables visual feedback of objects that are in collision or out of bounds and simulation ensures the physical plausibility of the virtual scenes. These capabilities are particularly important when choosing to move objects freely and deviating from their stable poses. It allows the design of complex scenes where objects are in contact with each other, e.g., piles and heaps.

We also provide a simple method to automatically generate collision-free and stable scene configurations.
This is done by randomly sampling $n$~objects from the library in one of their stable poses.
For each object, we choose a random translation in~x/y and rotation around~z and place it in the scene if it does not collide with already placed objects.
If no collision-free pose is found after $k$~tries, the object is not added and a scene with fewer than $n$~objects is returned.

\begin{table}[]
\centering
\caption{Overview of the most important functions of the SetupTool.}
\vspace{-0.5em}
\begin{tabular}{l|p{0.2\textwidth}}
\textbf{Function}                                            & \textbf{Description}                                                                                                                             \\ \hline
\multicolumn{1}{|l|}{\textit{\textbf{Load Object Library}}}  & \multicolumn{1}{l|}{load object library, start with empty scene}                                                                              \\ \hline
\multicolumn{1}{|l|}{\textit{\textbf{Load Scene}}}           & \multicolumn{1}{l|}{load an existing scene}                                                                                                      \\ \hline
\multicolumn{1}{|l|}{\textit{\textbf{Size}}}                 & \multicolumn{1}{l|}{choose size of the workspace/\aruco board}                                                                                    \\ \hline
\multicolumn{1}{|l|}{\textit{\textbf{Add Object}}}           & \multicolumn{1}{l|}{adds an object to the scene}                                                                                                 \\ \hline
\multicolumn{1}{|l|}{\textit{\textbf{Stable Poses}}}         & \multicolumn{1}{l|}{applies a stable pose to the selected object}                                                                       \\ \hline
\multicolumn{1}{|l|}{\textit{\textbf{Restrict Movement}}}    & \multicolumn{1}{l|}{restrict moving and rotating to xy-plane}                                                                                    \\ \hline
\multicolumn{1}{|l|}{\textit{\textbf{Validate \& Simulate}}} & \multicolumn{1}{l|}{\begin{tabular}[c]{@{}l@{}} collision and out-of-bounds checking \end{tabular}} \\ \hline
\multicolumn{1}{|l|}{\textit{\textbf{Display}}}              & \multicolumn{1}{l|}{switch between random and status colour}                                                                        \\ \hline
\multicolumn{1}{|l|}{\textit{\textbf{Random Configuration}}} & \multicolumn{1}{l|}{samples a random scene}                                                                            \\ \hline
\multicolumn{1}{|l|}{\textit{\textbf{Clear Scene}}}          & \multicolumn{1}{l|}{empties the scene}                                                                                                          \\ \hline
\multicolumn{1}{|l|}{\textit{\textbf{Save Scene}}}           & \multicolumn{1}{l|}{saves current scene to YAML file}                                                                                                 \\ \hline
\multicolumn{1}{|l|}{\textit{\textbf{Printout Page Size}}}        & \multicolumn{1}{l|}{sets the desired page size for the printout}                                                                               \\ \hline
\multicolumn{1}{|l|}{\textit{\textbf{Save Printout}}}        & \multicolumn{1}{l|}{saves printout of the current scene as PDF}                                                                               \\ \hline
\end{tabular}
\label{tab:setuptool-functions}
\vspaceafterfigure
\end{table}

\subsection{Training Data Generation}
The objects and scenes can be used to generate training data for data-driven grasping methods. This includes the rendering of training images (depth, RGB, and segmentation masks) as well as the sampling of grasp candidates. We generate antipodal grasp candidates by randomly sampling points on the surface of the mesh model as potential first contact points. We then cast rays within the friction cone of these points to identify potential second contact points where the rays intersect with the mesh surface. We filter the contact point pairs to ensure that the antipodal constraint is satisfied and sample various approach angles for the remaining pairs. Finally, a simplified gripper model is used to detect collisions and remove colliding grasps (see Fig.~\ref{fig:workflow_example}b).

\subsection{Grasp Simulation}
Our simulation environment is based on PyBullet~\cite{PyBullet} and used to: (1) validate the pre-computed stable poses of the objects in the object library; (2) ensure that scenes are physically plausible and that all objects are in a stable resting position during scene creation; and (3) simulate grasping trials. A Franka Panda gripper is loaded in the scene in the grasp pose. If it collides with any object in this pose, the grasp is considered to be unsuccessful; if not, the fingers are closed until contact is made and the gripper is lifted by $30cm$. If both fingers are still in contact with the target object, the grasp is considered successful. If friction parameters are included in the object library, they are used; otherwise, we use a friction coefficient of $0.24$. All objects are modeled with uniform density.

\subsection{Scene Arrangement for Real Experiments}
We employ two complementary strategies to facilitate the arrangement of objects in the real world. First, we create printouts similar to GRASPA templates~\cite{GRASPA_2020}, but instead of object silhouettes we generate more descriptive height maps.
We project the triangles of the object's mesh onto the ground plane determining its colour by the average height of each triangle's vertices. The closer the triangle is to the ground, the darker it will appear. This enables us to accurately reflect the orientation of a mug by indicating the handle position and it helps to visualise objects that overlap each other. However, the projections on the printout are inherently limited  when it comes to more complex scenes, e.g. stacked objects.

Our second strategy introduces a SceneVisualiser based on Augmented Reality to visualise objects in the real scene. The camera pose is estimated by recognising the \aruco marker board on the printout. An RGB image of the scene is rendered and blended with the real camera image. This helps to adjust the poses of real objects until they align with the rendered objects and imposes no limitations for stacked scenes.

\begin{figure}[tb]
    \centering
  	\begin{subfigure}[c]{0.48\columnwidth}
       \centering
       \includegraphics[width=\textwidth]{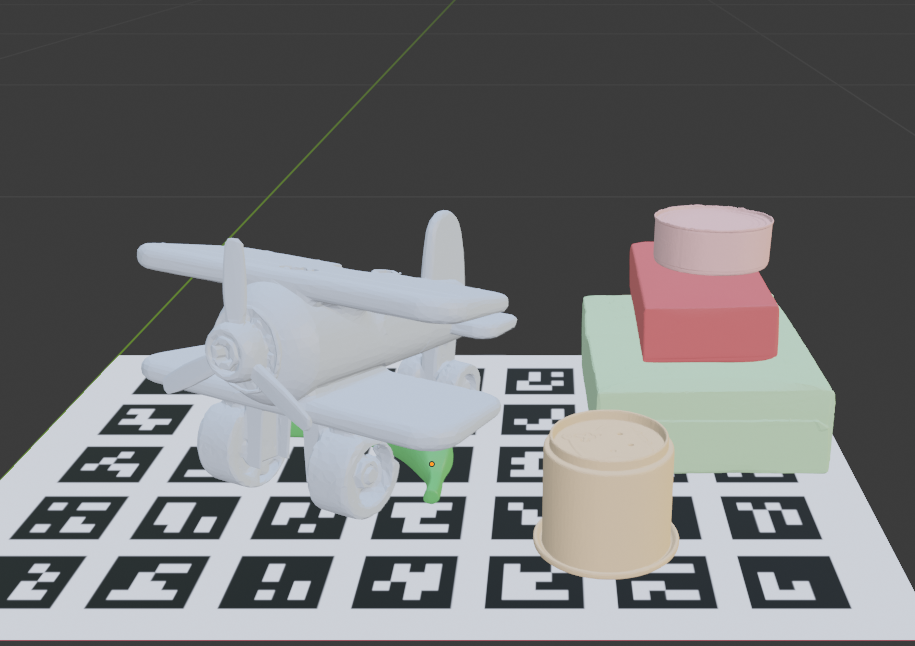}
  	\end{subfigure}
  	\begin{subfigure}[c]{0.48\columnwidth}
       \centering
       \frame{\includegraphics[width=\textwidth]{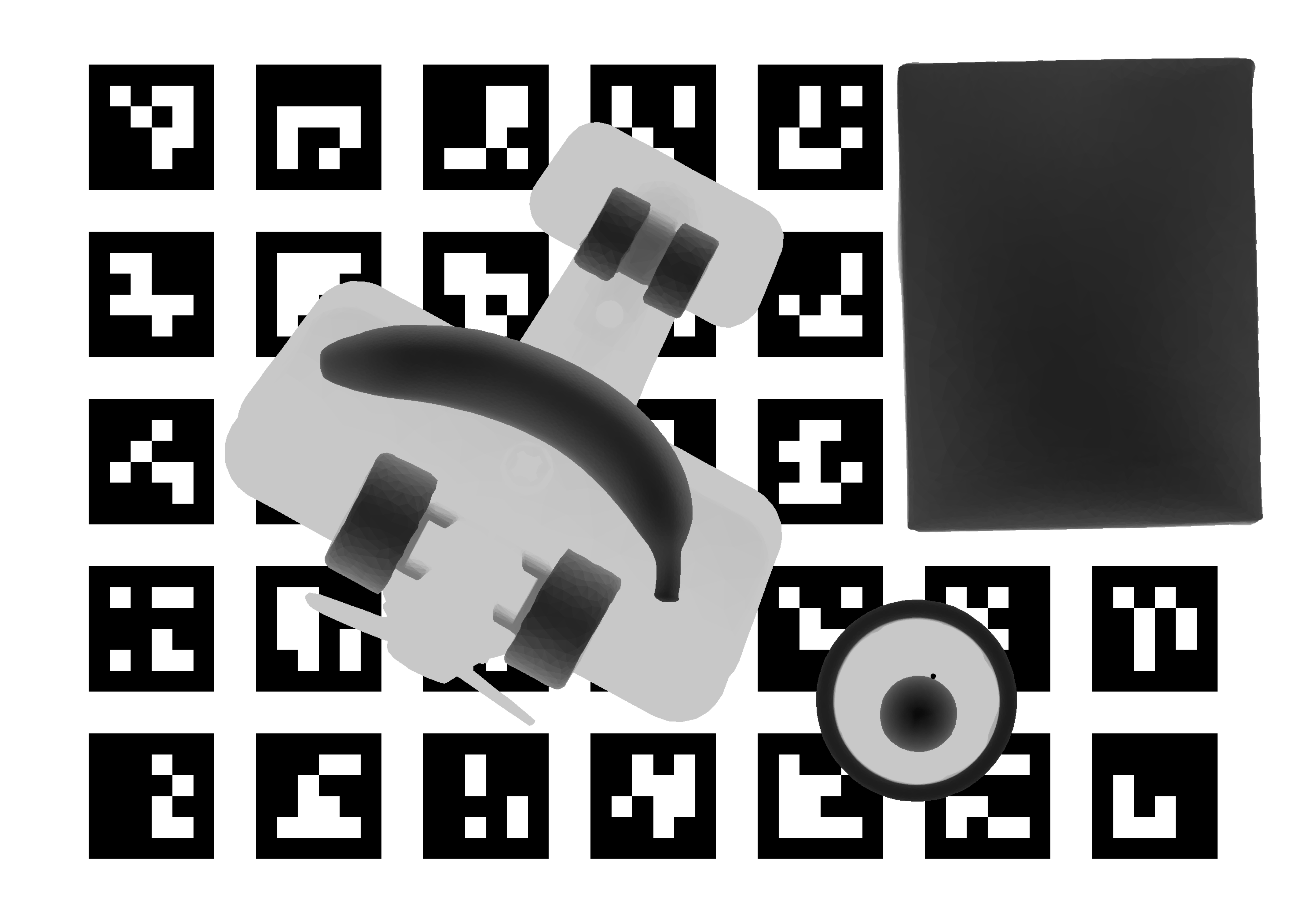}}
  	\end{subfigure}
  	\caption{Interesting challenges for robot grasping: 
  	    The banana under the airplane and the marble in the cup are perfectly visible on the printout (right).
  	    The SetupTool (left) also allows to create physically plausible piles of objects. The bottom item can be arranged using the printout; for others, the Augmented Reality SceneVisualiser is required.}
    \vspaceafterfigure
    \vspace{2mm}
  	\label{fig:interesting_use_cases}
\end{figure}

\begin{figure}[tb]
    \centering
  	\begin{subfigure}[c]{0.48\columnwidth}
       \centering
       \includegraphics[width=\textwidth]{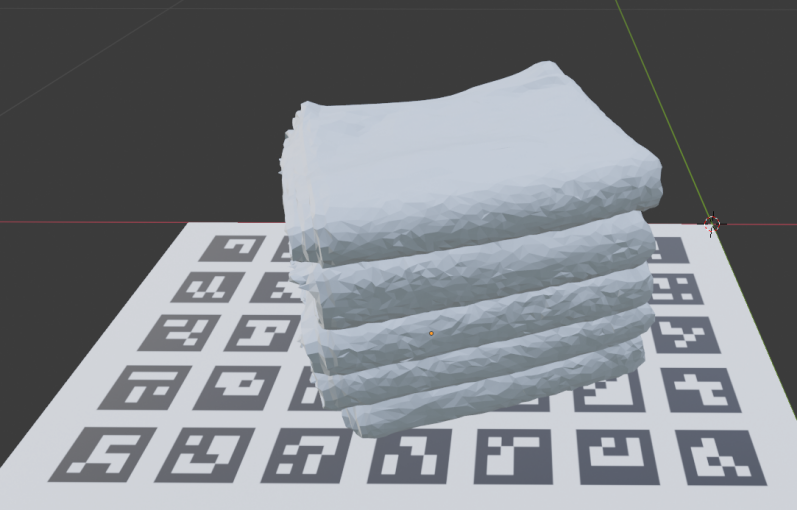}
  	\end{subfigure}
  	\begin{subfigure}[c]{0.48\columnwidth}
       \centering
       \includegraphics[width=\textwidth]{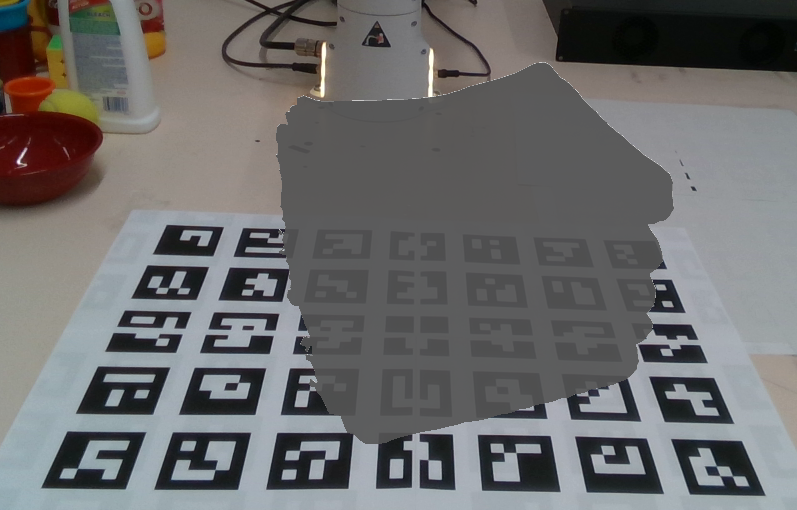}
  	\end{subfigure}
  	\caption{Our approach to scene creation and setup can be extended to deformable objects by treating a particular configuration as rigid. SetupTool (left) and SceneVisualiser (right).}
    \vspaceafterfigure
  	\label{fig:towel}
\end{figure}

Although the two strategies can be applied independently, we found that their combination is best for fast and accurate placements. Fig.~\ref{fig:interesting_use_cases} shows some interesting scene compositions that can be accomplished with our method. We can also extend our approach to deformable objects. The simulator can not accurately model these objects, but given a mesh with the desired configuration of the deformable object, e.g., from a 3D scanner, we can treat it as a rigid object and place it in the scene, create printouts, and use the SceneVisualiser to set up scenes for robot experiments (see Fig.~\ref{fig:towel}). When used with benchmark datasets such as the Household Cloth Object Set~\cite{ClothObjectSet_2022}, this can boost reproducibility for manipulation of deformable objects.

%% file: 03-experiments.tex
\begin{figure}[tb]
    \centering
  	\begin{subfigure}[c]{0.48\columnwidth}
       \centering
       \includegraphics[width=\textwidth]{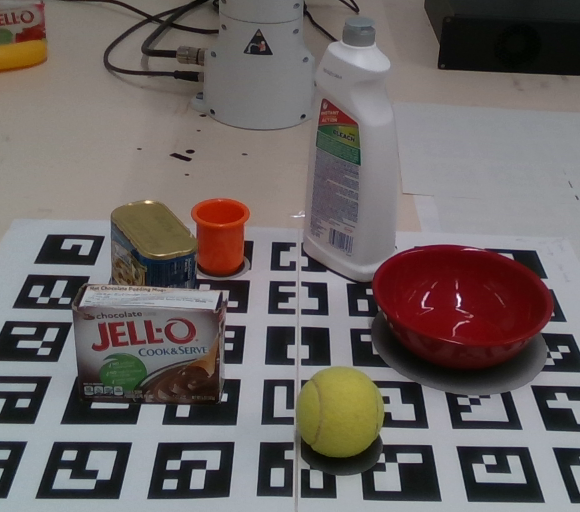}
  	\end{subfigure}
  	\begin{subfigure}[c]{0.48\columnwidth}
       \centering
       \includegraphics[width=\textwidth]{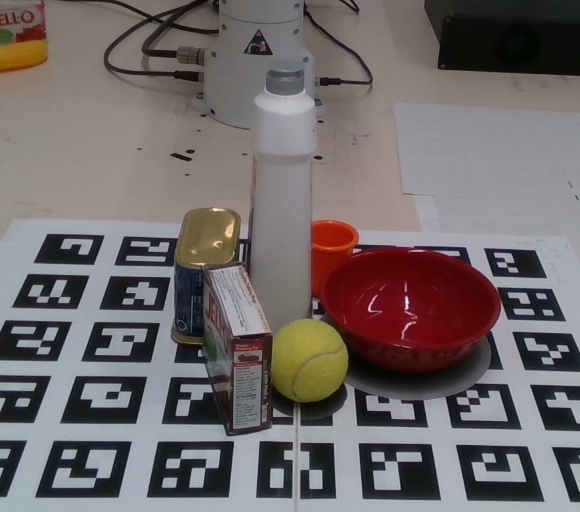}
  	\end{subfigure}
  	\caption{The two scenes we used for our experiments.}
    \vspaceafterfigure
    \vspace{2mm}
  	\label{fig:experiment-scenes}
\end{figure}

The BURG-Toolkit can be used to analyse the sim-to-real gap in robot grasping, which we demonstrate with a proof-of-concept experiment. We created two different scenes with the same six YCB objects (see Fig.~\ref{fig:experiment-scenes}).
In the first, all objects are relatively wide apart whereas in the second scene the objects are much closer, almost touching each other, increasing the importance of collision and contact modeling.
Using the BURG-Toolkit, we sample 1000~grasps for each object in each scene, filter grasps that fail simple collision checking, and simulate the remaining candidates.
From this set, we randomly select 10~grasps per object based on the simulation result; we try to select a balanced set of successful and unsuccessful ones. This gives us 60~grasps per scene which we execute with a real robot. Similar to the simulation experiments, if the robot grasps and lifts the target object by $30cm$, the grasp is considered successful.
We use a Franka Emika Panda robot with the Panda hand.
The offset between robot base and scene, the object library and scene files as well as the grasps are available on our project website to allow reproduction of our experiments.

\begin{figure}[tb]
    \footnotesize
    \centering
  	\begin{subfigure}[t]{0.48\columnwidth}
       \centering
        \begin{tabular}{l|c|c|c|c}
             \multicolumn{2}{c}{} & \multicolumn{2}{c}{real} & \\ \cline{3-4}
             \multicolumn{1}{c}{} & & 0 & 1 & \multicolumn{1}{c}{$\Sigma$} \\ \cline{2-4}
             \multirow{2}{*}{sim} & 0 &     20  & 10 & 30   \\ \cline{2-4}
                                  & 1 &     9   & 21 & 30   \\ \cline{2-4}
             \multicolumn{1}{c}{} & 
                    \multicolumn{1}{c}{$\Sigma$} & 
                    \multicolumn{1}{c}{29} & 
                    \multicolumn{1}{c}{31} & 60 \\
            
            \multicolumn{5}{c}{} \\
            \multicolumn{3}{r}{Precision:} & \multicolumn{2}{c}{$70.00\%$} \\
            \multicolumn{3}{r}{Recall:} & \multicolumn{2}{c}{$67.74\%$} \\
        \end{tabular}
        \caption{Scene with objects wide apart.}
  	\end{subfigure}
  	\begin{subfigure}[t]{0.48\columnwidth}
       \centering
        \begin{tabular}{l|c|c|c|c}
             \multicolumn{2}{c}{} & \multicolumn{2}{c}{real} & \\ \cline{3-4}
             \multicolumn{1}{c}{} & & 0 & 1 & \multicolumn{1}{c}{$\Sigma$} \\ \cline{2-4}
             \multirow{2}{*}{sim} & 0 &     25  & 12 & 37   \\ \cline{2-4}
                                  & 1 &     13  & 10 & 23   \\ \cline{2-4}
             \multicolumn{1}{c}{} & 
                    \multicolumn{1}{c}{$\Sigma$} & 
                    \multicolumn{1}{c}{38} & 
                    \multicolumn{1}{c}{22} & 60 \\
            
            \multicolumn{5}{c}{} \\
            \multicolumn{3}{r}{Precision:} & \multicolumn{2}{c}{$43.48\%$} \\
            \multicolumn{3}{r}{Recall:} & \multicolumn{2}{c}{$45.45\%$} \\
        \end{tabular}
        \caption{Scene with objects close together.}
  	\end{subfigure}
    \caption{Confusion matrices, precision and recall of the simulation, considering the real-world grasp outcomes as ground truth. A failed grasp is indicated by~0, a successful grasp by~1.
    }
  	\vspaceafterfigure
    \label{fig:experiments}
\end{figure}

For quantification, we consider the simulator as a binary classifier and the outcomes from real-world experiments as ground truth. The results in Fig.~\ref{fig:experiments} show that there is a substantial difference in classification performance between the two scenes, with more errors observed in the second scene with objects close together. Interestingly, for each of the two scenes, the precision and recall values are similar. This indicates that the simulation does not systematically over- or under-estimate the real-world grasping performance. Typical errors correspond to situations in which the simulation reports a successful grasp but no trajectory can be found with the robot arm in the physical world, or situations in which the physical gripper moves objects away while approaching the grasp pose, giving a successful grasp, but the grasp fails in simulation because a collision is detected. Although we only evaluated two scenes, these results do indicate a large sim-to-real gap for robot grasping, particularly in scenes where objects are closer to each other. We will explore this further in future experiments.

%% file: 04-conclusion.tex
We have designed our BURG-Toolkit to facilitate benchmarking of methods and reproducibility of experimental results in robot grasping, and to investigate the sim-to-real gap. In the future, we will focus on benchmarking different grasping methods as well as incorporating additional measures of grasp success (i.e., other than simple success or failure of grasps) to better quantify the sim-to-real gap. In the long-term, we are committed to further developing these tools and appreciate feedback by the research community.

